%% file: main.tex
\documentclass[10pt,conference]{IEEEtran}

\usepackage{cite}
\usepackage{amsmath,amssymb,amsfonts}
\usepackage{algorithmic}
\usepackage{graphicx}
\usepackage{textcomp}
\usepackage{xcolor}
\usepackage{array}
\usepackage{tikz}
\usepackage{pgfplots}
\usepackage{booktabs}
\usepackage{multirow}
\usepackage{url}
\usepackage{algorithm}
\usepackage{balance}
\pgfplotsset{compat=1.17}
\usepackage{url}
\usepackage{balance}
\usepackage{pifont}
\usepackage{comment}

\begin{document}

\title{Scalable Explainability-as-a-Service (XaaS) for Edge AI Systems}

\author{\IEEEauthorblockN{
1\textsuperscript{st} Samaresh Kumar Singh\,
}
\IEEEauthorblockA{\textit{
IEEE Senior Member} \\
Leander, Texas, USA \\
ssam3003@gmail.com}
\and
\IEEEauthorblockN{
2\textsuperscript{st} Joyjit Roy\,
}
\IEEEauthorblockA{\textit{
IEEE Senior Member} \\
Austin, Texas, USA \\
joyjit.roy.tech@gmail.com}
}

\maketitle

\input{abstract}
\input{introduction}
\input{related_work}

\input{problem_formulation}
\input{architecture}
\input{method}
\input{experimental_setup}
\input{results}
\input{discussion}
\input{limitations}
\input{conclusion}

\input{future_work}

\bibliographystyle{IEEEtran}
\bibliography{references}

\end{document}

%% file: abstract.tex
\begin{abstract}
Though Explainable AI (XAI) has made significant advancements, its inclusion in edge and IoT systems is typically ad-hoc and inefficient. Most current methods are "coupled" in such a way that they generate explanations simultaneously with model inferences. As a result, these approaches incur redundant computation, high latency and poor scalability when deployed across heterogeneous sets of edge devices. In this work we propose \textit{Explainability-as-a-Service} (\textbf{XaaS}), a distributed architecture for treating explainability as a first-class system service (as opposed to a model-specific feature). The key innovation in our proposed \textit{XaaS} architecture is that it decouples inference from explanation generation allowing edge devices to request, cache and verify explanations subject to resource and latency constraints. To achieve this, we introduce three main innovations: (1) A distributed explanation cache with a semantic similarity based explanation retrieval method which significantly reduces redundant computation; (2) A lightweight verification protocol that ensures the fidelity of both cached and newly generated explanations; and (3) An adaptive explanation engine that chooses explanation methods based upon device capability and user requirement. We evaluated the performance of \textbf{XaaS} on three real-world edge-AI use cases: (i) manufacturing quality control; (ii) autonomous vehicle perception; and (iii) healthcare diagnostics. Experimental results show that XaaS reduces latency by 38\% while maintaining high explanation quality across three real-world deployments. Overall, this work enables the deployment of transparent and accountable AI across large scale, heterogeneous IoT systems, and bridges the gap between XAI research and edge-practicality.
\end{abstract}


%% file: introduction.tex
\section{Introduction}

Edge AI is everywhere now, but most systems cannot explain their decisions. The use of AI on edge devices has grown rapidly in recent years \cite{zhou2019edge}. However, the integration of explainable AI is a major challenge. Although XAI techniques such as LIME \cite{ribeiro2016should}, SHAP \cite{lundberg2017unified} and GradCAM \cite{selvaraju2017grad} can make models transparent, they incur high computational costs (over 1000 model evaluations), which makes them unsuitable for edge devices with limited resources.

Three types of inefficiencies exist today: \textbf{(1)} \emph{Redundant computation}: Edge devices generate explanations independently for each input that is received (similar inputs); \textbf{(2)} \emph{Resource mismatch}: The XAI methods developed for centralized environments cannot be adapted to support a variety of edge hardware configurations; \textbf{(3)} \emph{No reuse}: Each time an explanation is generated it is discarded and will have to be regenerated again for the next query of a similar nature.

This paper presents \textit{Explainability-as-a-Service} (\textit{XaaS}), which treats the explanation process as an additional distributed service that includes caching, verification, and adaptive delivery of explanations to devices. 

%% file: related_work.tex
\section{Related Work}
\label{sec:related}

We group other research under four categories: Explainable AI Methods, Edge-AI Systems, XAI in Limited Resource Environments, Caching and Service Architectures.

\subsection{Explainable AI Methods}

Local Explanation Methods Model-agnostic XAI techniques include SHAP \cite{lundberg2017unified}, LIME \cite{ribeiro2016should}, and both use local linear approximation models to explain an output of a model. Both have limitations due to the number of model calls required to be made ($>$ 1000). This severely limits their potential for deployment on Edge platforms.

Gradient-based methods are efficient but model-specific, and can also produce noise. Counterfactual Explanations identify a minimal change to the input that will result in an alternative prediction and provide actionable knowledge. The cost of optimizing per instance is high when generating large numbers of explanations.

Prior research has examined the quality and generality of explanations. No prior study has explored the system-wide aspects of caching/reuse/delivering content to a variety of devices.

\subsection{Edge AI and Low Latency Inference}
Edge AI enables local inference through model compression
techniques such as MobileNets \cite{howard2017mobilenets}, EfficientNets \cite{tan2019efficientnet}, and SqueezeNet \cite{iandola2016squeezenet}, along with inference partitioning across the edge-cloud continuum \cite{teerapittayanon2017distributed} . Federated learning further supports training distributed AI models across multiple data sources. While both techniques decrease inference latency, neither technique offers a method for providing explanations for an AI model's output. XaaS addresses the need for explanation services in the area of AI model deployment by providing explanation capabilities as an additional service layer over existing AI models.

\subsection{XAI For Resource-Constrained Devices}

Srinivasan et al. \cite{srinivasan2021explainable} proposed low-cost saliency techniques for CNN-based edge vision models. This reduced computational cost by approximately 60\% through approximations but was limited to CNNs without caching mechanisms. Islam et al. \cite{islam2021edge} used knowledge distillation to generate small-sized explainer networks for edge deployment. This required training new explainer networks and did not generalize across explanation methods.

Bhatt et al. \cite{bhatt2020explainable} identified challenges in applying XAI to Edge Computing and advocated for adaptable, resource-sensitive explanation methods, a void that XaaS fills.

All previous approaches were model-based and did not offer a framework for creating, storing, or reusing XAI at the system level.

\subsection{Caching and Service Architecture}

Caching is essential for distributed systems. Content Delivery Networks (CDNs) \cite{nygren2010akamai} cache web pages to deliver them quickly, and HTTP mechanisms like cache-control headers and ETags \cite{fielding1999rfc2616} handle cache validity.

While there are many prediction services available, none of these have taken into account explainability. Bhatt et al. \cite{bhatt2020machine} had proposed the idea of machine learning-as-a-service (MLaaS) that would include an explanation component. However, there was no real-world architectural design. Meske and Bunde \cite{meske2020transparency}, presented a "transparency-as-a-service" concept based upon regulatory compliance, but this was done with no real-world technical solution.

Applying caching to explanations is complex. They are high-dimensional objects whose quality depends on model state and input similarity, with varying fidelity requirements.

\subsection{Summary and Differentiation}

Table \ref{tab:related_comparison} provides an overview of how XaaS compares to prior related works in many key areas.

\begin{table*}[t]
\vspace{0.05in}
\centering
\caption{Comparison of XaaS with Related Work}
\label{tab:related_comparison}
\begin{tabular}{p{3.5cm}ccccccc}
\toprule
\textbf{Work} & \textbf{XAI Methods} & \textbf{Edge/IoT} & \textbf{Caching} & \textbf{Verification} & \textbf{Adaptive} & \textbf{Heterogeneity} & \textbf{System Eval} \\
\midrule
LIME \cite{ribeiro2016should} & \checkmark & $\times$ & $\times$ & $\times$ & $\times$ & $\times$ & $\times$ \\
SHAP \cite{lundberg2017unified} & \checkmark & $\times$ & $\times$ & $\times$ & $\times$ & $\times$ & $\times$ \\
GradCAM \cite{selvaraju2017grad} & \checkmark & $\times$ & $\times$ & $\times$ & $\times$ & $\times$ & $\times$ \\
Edge AI Systems \cite{zhou2019edge} & $\times$ & \checkmark & $\times$ & $\times$ & Partial & \checkmark & \checkmark \\
Srinivasan et al. \cite{srinivasan2021explainable} & \checkmark & \checkmark & $\times$ & $\times$ & $\times$ & $\times$ & Partial \\
Islam et al. \cite{islam2021edge} & \checkmark & \checkmark & $\times$ & $\times$ & $\times$ & $\times$ & Partial \\
CDNs \cite{nygren2010akamai} & $\times$ & Partial & \checkmark & \checkmark & \checkmark & \checkmark & \checkmark \\
ML Serving \cite{olston2017tensorflow} & $\times$ & Partial & Partial & $\times$ & \checkmark & \checkmark & \checkmark \\
\midrule
\textbf{XaaS} & \checkmark & \checkmark & \checkmark & \checkmark & \checkmark & \checkmark & \checkmark \\
\bottomrule
\end{tabular}
\end{table*}

XaaS represents the first architecture to integrate XAI and Edge Computing to provide scalable explanations across multiple IoT deployment types. Prior work lacks complete implementation of XAI at the Edge using all four features (caching, verification, adaptive delivery) or the wide range of evaluations based on real-world use cases presented in this paper.

%% file: problem_formulation.tex
\section{Problem Formulation}
\label{sec:problem}

We define the Edge Explainability Problem by formalizing the System Model, Objectives, and Constraints that an XaaS Solution must satisfy.

\subsection{System Model}

An Edge AI Deployment with $N$ Devices $\mathcal{D} = \left\{d_{1}, d_{2}, \ldots, d_{N}\right\}$, each Device $d_i$ has following properties:

\begin{itemize}
\item \textbf{Model} $f_i: \mathcal{X} \to \mathcal{Y}$: The mapping from Input to Prediction.
\item \textbf{Compute Capacity} $C_i$: The processing power available for computations at Device $i$ (FLOPS, Cores, Memory).
\item \textbf{Network} $B_i$: The bandwidth to edge servers or cloud for communication purposes at Device $i$.
\item \textbf{Latency Tolerance} $L_i$: The maximum time allowed to wait for the explanation in Device $i$. 
\end{itemize}

The system also has $M$ Edge Servers, $\mathcal{S} = \{s_1, s_2, ..., s_M\}$, with much greater computational capabilities for generating explanations and maintaining cache.

At time $t$, device $d_i$ receives input $\mathbf{x}_t \in \mathcal{X}$ and produces prediction $\hat{y}_t = f_i(\mathbf{x}_t)$. A user or downstream System can request an explanation $E(\mathbf{x}_t, f_i)$ as to why $f_i$ produced $\hat{y}_t$.

\subsection{Explanation Methods}

Let $\mathcal{M} = \{m_1,m_2,\cdots,m_K\}$ be the set of possible methods for generating explanations (LIME, SHAP, GradCAM, etc.) for each explanation method $m_k$:

\begin{itemize}
\item \textbf{Computational cost}, $\text{cost}(m_k,d_i)$: Energy required to generate an explanation using $m_k$ on a device $d_i$.
\item \textbf{Fidelity}, $\text{fid}(m_k,f_i)$: A measure of how well $m_k$ explains $f_i$, which we define here as $1-\frac{1}{N}\sum_i|f(\mathbf{x}_i)-g(\mathbf{x}_i)|$, where $g$ is the explanation model.
\item \textbf{Applicability}, $\text{app}(m_k,f_i)$: If $m_k$ can provide an explanation for $f_i$. For example, if $f_i$ is a CNN, then $\text{app}(GradCAM,f_i)=1$, but $\text{app}(LIME,f_i)=0$. 
\end{itemize}

The cost function will include both the cost of generating a given explanation and communication costs (if the explanation is generated in a remote environment).

\begin{equation}
\text{cost}(m_k, d_i) = \alpha \cdot T_{\text{compute}}(m_k, C_i) + \beta \cdot T_{\text{comm}}(m_k, B_i)
\label{eq:cost}
\end{equation}

Where $\alpha,\beta$ represent weights for computation vs. communication depending on application priorities (latency, power consumption, etc.).

\subsection{Explanation Requests}

Over time horizon $[0, T]$, the system receives explanation requests $\mathcal{R} = \{r_1, r_2, \ldots, r_Q\}$. Request $r_j$ specifies:

\begin{itemize}
    \item \textbf{Input} $\mathbf{x}_j$: The instance to explain
    \item \textbf{Model} $f_i$: Which model's prediction to explain
    \item \textbf{Device} $d_i$: Where the request originates
    \item \textbf{Requirements} $\rho_j$: Fidelity threshold $\rho_j^{\text{fid}}$, latency bound $\rho_j^{\text{lat}}$
\end{itemize}

\subsection{Naive Approach and Its Limitations}

The naive approach generates each explanation independently:

\begin{equation}
E(\mathbf{x}_j, f_i) = m_k(\mathbf{x}_j, f_i) \quad \forall r_j \in \mathcal{R}
\label{eq:naive}
\end{equation}

where $m_k$ is selected based on device capabilities.

This method has three major disadvantages in terms of efficiency:

\textbf{(1) Redundancy in Computation:} A similar input leads to a redundant generation of explanations. In a formal way, if we have that $\mathbf{x}_j \approx \mathbf{x}_{j'}$ (similar input), and also $f_i(\mathbf{x}_j)=f_i(\mathbf{x}_{j'})$, then $E(\mathbf{x}_j,f_i)\approx E(\mathbf{x}_{j'},f_i)$ due to the explanation consistency (\cite{arrieta2020explainable}). But these explanations are calculated separately, which wastes resources.

\textbf{(2) Sub-Optimal Choice of Method:} The device $d_i$ might be insufficiently capable to perform methods with high fidelity; therefore, the explanations produced on this device will be of lower quality than those which can be made from an edge server, which is close to it ($s \in \mathcal{S})$ but can produce good explanations at latency boundaries.

\textbf{(3) Lack of Verification:} Explanations produced before updating the model (for example, using Federated Learning or Retraining) may no longer be valid. Therefore, there is a risk that outdated explanations will mislead the user without a verification process.

\subsection{XaaS Objectives}

XaaS intends to reduce overall system costs while meeting both explanation quality and latency standards by:

\begin{align}
\min \quad & \sum_{r_j \in \mathcal{R}} \text{cost}(r_j) \label{eq:objective} \\
\text{s.t.} \quad & \text{fidelity}(E(\mathbf{x}_j, f_i)) \geq \rho_j^{\text{fid}} \quad \forall r_j \label{eq:fidelity_constraint} \\
& \text{latency}(r_j) \leq \rho_j^{\text{lat}} \quad \forall r_j \label{eq:latency_constraint} \\
& E(\mathbf{x}_j, f_i) \text{ valid for model version } v_i \label{eq:validity_constraint}
\end{align}

where $\text{cost}(r_j)$ represents the cost of processing the request $r_j$, including computation and communication.

The reason XaaS can cut costs is through \textit{reusing explanations}. If we can determine that $E(\mathbf{x}_j, f_i) \approx E(\mathbf{x}_{j'}, f_i)$, then it is possible to use the cached explanation again without having to regenerate it and still maintain explanation fidelity.

\subsection{Caching Validity Conditions}

For explanation caching to be sound, there must be conditions under which a cached explanation remains valid. Let $E_{\text{cache}}(\mathbf{x}',  f_i, v)$ represent an explanation for $\mathbf{x}'$, model $f_i$, version $v$ that was previously cached.

\textbf{Condition 1 (Semantic Similarity)}: The new input $\mathbf{x}$ should be semantically similar to the input that was previously stored in memory:

\begin{equation}
d_{\text{sem}}(\mathbf{x}, \mathbf{x}') < \epsilon_{\text{sim}}
\label{eq:similarity_condition}
\end{equation}

Where $d_{\text{sem}}$ is a domain-appropriate similarity metric (e.g., perceptual distance for images, cosine similarity for embeddings) and $\epsilon_{\text{sim}}$ is a threshold.

\textbf{Condition 2 (Prediction Consistency)}: Both inputs should produce the same prediction:

\begin{equation}
f_i(\mathbf{x}) = f_i(\mathbf{x}')
\label{eq:prediction_condition}
\end{equation}

This ensures the target of the prediction or explanation remains relevant.

\textbf{Condition 3 (Model Version)}: The cached explanation should exist for the current version of the model:

\begin{equation}
v = v_{\text{current}}
\label{eq:version_condition}
\end{equation}

or confirmed as still valid (see Section~\ref{sec:verification}).

\textbf{Condition 4 (Fidelity Preservation)}: The cached explanation must meet the request for the fidelity requirement:

\begin{equation}
\text{fidelity}(E_{\text{cache}}(\mathbf{x}', f_i, v), \mathbf{x}, f_i) \geq \rho_j^{\text{fid}}
\label{eq:fidelity_preservation}
\end{equation}

When these four conditions hold, XaaS can safely retrieve the cached explanation, reducing cost from $\text{cost}(m_k,d_i)$ to $\text{cost}_{\text{cache}}(d_i,s)$ (2-3 orders of magnitude lower).

\subsection{Optimization Challenges}
There are several reasons why the optimization problem of equations \eqref{eq:objective}-\eqref{eq:validity_constraint} is difficult and it is hard to solve for several reasons. There are many dimensions of the decision space (cache vs generate, methods to use, where to do the computation, how to rank candidates for every input). All of device capability, network condition, cache content change with time. There are also costs associated with similarity assessment; therefore, it has to weigh the cost of looking up something versus the cost/benefit of storing in cache. Also, this is an implicit multi-objective optimization problem (latency, energy, fidelity, and cache hit rate) and depending on the application they can weight the objectives very differently.

We present, from Section~\ref{sec:architecture} to Section~\ref{sec:adaptive}, practical methods to approximate the solution to this optimization problem using caching heuristics, lightweight verification, and adaptive method selection.

\subsection{Assumptions}

Our formulation relies on several key assumptions as follows:

\textbf{A1 (Consistent Explanations):} We assume that if two inputs are the same, then their predictions will be the same; therefore, they should have the same explanation. A consistent explanation is an expected property of many explanation methods in XAI \cite{arrieta2020explainable}. So, we would expect this to be true for most cases.

\textbf{A2 (Structured Input Distribution):} The input data distribution is structured (for example, images or sensor data) as opposed to being randomly generated. This provides us with the opportunity to create an organized cache based on similarity between input data.

\textbf{A3 (Tolerant Latency):} An application can tolerate milliseconds to seconds of delay in providing explanations. This is reasonable when explaining results to humans, but would likely not be acceptable for real-time control loop applications.

\textbf{A4 (Stability of Model):} In comparison to the frequency of explanation requests, models update at a relatively slow rate (i.e., hours to days). This allows our method to amortize over long periods of time.

\textbf{A5 (Communication Available):} Devices can send data to an edge server; however, communication quality varies by device and connection. Communication availability is an assumed condition in edge computing.

The assumptions listed above provide a basis for the decision-making process in the design of XaaS.

%% file: architecture.tex
\section{XaaS Architecture}
\label{sec:architecture}

Figure~\ref{fig:architecture} shows the XaaS architecture with five components:

\begin{figure*}[t]
\centering
\includegraphics[width=0.95\textwidth]{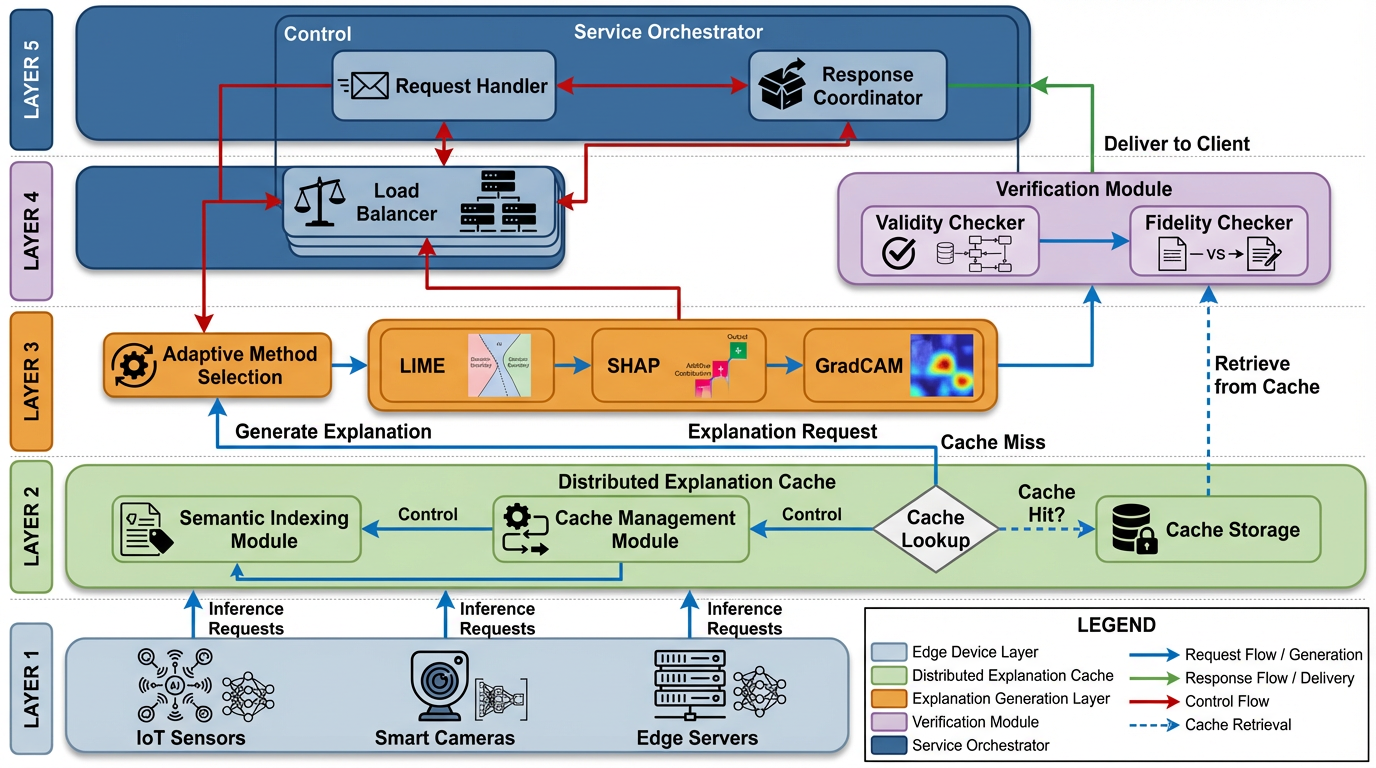}
\caption{XaaS System Architecture. The framework decouples inference from explanation generation, enabling edge devices to request, cache, and verify explanations efficiently.}
\label{fig:architecture}
\end{figure*}

\textbf{(1) Edge Device Layer:} Heterogeneous devices run AI models and request explanations via the XaaS service.

\textbf{(2) Distributed Explanation Cache:} Two-tier hierarchical cache with local caches on edge servers for low-latency access (5-10ms) and a global cloud cache for comprehensive coverage (50-100ms). Uses FAISS \cite{johnson2019billion} for nearest-neighbor similarity search over embeddings.

\textbf{(3) Explanation Generation Layer:} Generates explanations on cache misses using LIME, SHAP, GradCAM, or attention mechanisms. Selects methods adaptively based on device capabilities.

\textbf{(4) Verification Module:} Ensures cached explanations remain valid via lightweight protocols checking model version and fidelity.

\textbf{(5) Service Orchestrator:} Coordinates request routing, load balancing, and model version management.

\textbf{Request Flow:} Device submits request, orchestrator queries cache using embedding $\mathbf{e} = \phi(\mathbf{x})$ (CLIP \cite{radford2021learning} for images, BERT for text). On cache hit, verify and return. On cache miss, generate, cache, and return.

%% file: method.tex
\section{Core Algorithms}
\label{sec:method}

\subsection{Semantic Explanation Caching}
\label{sec:caching}

Utilizing the semantic caching concept, Algorithm \ref{alg:cache_lookup} uses semantic similarity with the FAISS \cite{johnson2019billion} library to perform efficient nearest neighbor search of cached explanations. The embeddings, denoted as $\phi(\mathbf{x})$, are calculated using specific domain models: CLIP \cite{radford2021learning} for image inputs and either BERT or Sentence-BERT for text inputs.

\begin{algorithm}[h]
\caption{Cache Lookup}
\label{alg:cache_lookup}
\begin{algorithmic}[1]
\REQUIRE Query $(\mathbf{x}_q, f, \rho)$, Cache $C$, Threshold $\epsilon_{\text{sim}}$
\STATE $\mathbf{e}_q \leftarrow \phi(\mathbf{x}_q)$; $\text{candidates} \leftarrow \text{NN}_k(C, \mathbf{e}_q)$
\FOR{each $(E_i, \mathbf{e}_i, v_i, \text{fid}_i)$ in candidates}
    \IF{$\|\mathbf{e}_q - \mathbf{e}_i\|_2 > \epsilon_{\text{sim}}$}
        \STATE \textbf{continue}
    \ENDIF
    \IF{$f(\mathbf{x}_q) \neq \text{cached prediction}$}
        \STATE \textbf{continue}
    \ENDIF
    \IF{$v_i \neq v_{\text{current}}$ \AND \NOT Verify$(E_i, \mathbf{x}_q, f)$}
        \STATE \textbf{continue}
    \ENDIF
    \IF{$\text{fid}_i \geq \rho^{\text{fid}}$}
        \RETURN $E_i$
    \ENDIF
\ENDFOR
\RETURN NULL
\end{algorithmic}
\end{algorithm}

The similarity threshold, $\epsilon_{\text{sim}}$, is dynamically adjusted (online) to meet a trade-off between hit-rate and fidelity through the use of consistency-based measures of fidelity that have been previously documented in the XAI literature \cite{arrieta2020explainable}.

\subsection{Lightweight Verification}
\label{sec:verification}

As shown in Algorithm~\ref{alg:verification}, the number of perturbations in this case is $n=15$ when cached explanations are produced for model version $v \neq v_{\text{current}}$. In comparison, a full LIME explanation requires over 1000 perturbations \cite{ribeiro2016should}, while a full SHAP explanation requires thousands of additional perturbations \cite{lundberg2017unified}:

\begin{algorithm}[h]
\caption{Lightweight Verification}
\label{alg:verification}
\begin{algorithmic}[1]
\REQUIRE Cached $E_{\text{cache}}$, Query $\mathbf{x}_q$, Model $f$
\STATE Generate perturbations $\{\mathbf{x}_1, \ldots, \mathbf{x}_n\}$
\FOR{$i = 1$ to $n$}
    \STATE Compare $f(\mathbf{x}_i)$ with $E_{\text{cache}}.predict(\mathbf{x}_i)$
\ENDFOR
\STATE \textbf{return} fidelity $\geq$ threshold
\end{algorithmic}
\end{algorithm}

The approach also leverages perturbation-based fidelity assessment, which is able to detect 95.5\% of invalid explanations with a 3.2\% loss in performance.

\subsection{Adaptive Method Selection}
\label{sec:adaptive}

When there is a cache miss, choose a method $m^*$ and a location $\ell^*$ that minimizes overall latency, given the fidelity and capacity requirements. Methods available for selection are: LIME \cite{ribeiro2016should}, SHAP \cite{lundberg2017unified}, GradCAM \cite{selvaraju2017grad}, Integrated Gradients, and attention mechanisms \cite{vaswani2017attention}.

\begin{equation}
(m^*, \ell^*) = \arg\min_{m, \ell} T_{\text{compute}}(m, \ell) + T_{\text{comm}}(\ell, d)
\end{equation}

A greedy search strategy will first favor explanation methods with the highest fidelity values and then attempt different locations (edge, device, or cloud) until it finds a combination of an explanation generation method and an area that meets the feasibility criteria in  $O(|\mathcal{M}| \cdot |\mathcal{L}|)$ time complexity. The greedy strategy can be viewed as a form of partitioned inference strategy, modified to fit the explanation-generation process.

%% file: experimental_setup.tex
\section{Experimental Setup}
\label{sec:setup}

\textbf{Scenarios:} (1) \textit{Manufacturing Quality Control (MQC)}: In this case there were 150 devices, 127K sampling points and eight classes of defects. 
\\
(2) \textit{Autonomous Vehicle Fleet (AVF):}  In this case there were 80 vehicles, 215K scenarios to drive.
\\
(3) \textit{Healthcare Monitoring (HCM):} There were 200 patients, and 89K samples of their vital signs.

\textbf{Hardware:} Edge devices (Raspberry Pi 4B, NVIDIA Jetson Nano, Intel NUC), 5 edge servers per scenario (Tesla T4 GPU, 1000-entry cache), cloud server (10K global cache).

\textbf{Baselines:} \textit{LocalGen} (on-device generation), \textit{CloudXAI} (SHAP/LIME in the cloud), \textit{EdgeXAI} (edge generation with no caching), \textit{FedXAI} (federated XAI).

\textbf{Metrics:} End-to-end latency (ms), explanation fidelity ($\phi \geq 0.9$ target), cache hit rate (\%), Throughput (requests/sec.), Success Rate (\%).

\textbf{Configuration:} $\epsilon_{\text{sim}} \in [0.12, 0.18]$, $\phi_{\text{target}} = 0.92$, $n = 15$ verification samples. Each experiment had 2-hour warm-up, 24-hour run, five repetitions.

%% file: results.tex
\section{Results}
\label{sec:results}

XaaS demonstrates substantial performance improvements across the three use cases examined, with explanations generated by XaaS performing better than baseline methods; results shown are averages with 95\% confidence intervals from five separate experimental runs.

\subsection{Primary Performance Comparison}

\begin{figure*}[t]
    \centering
    \includegraphics[width=0.95\textwidth]{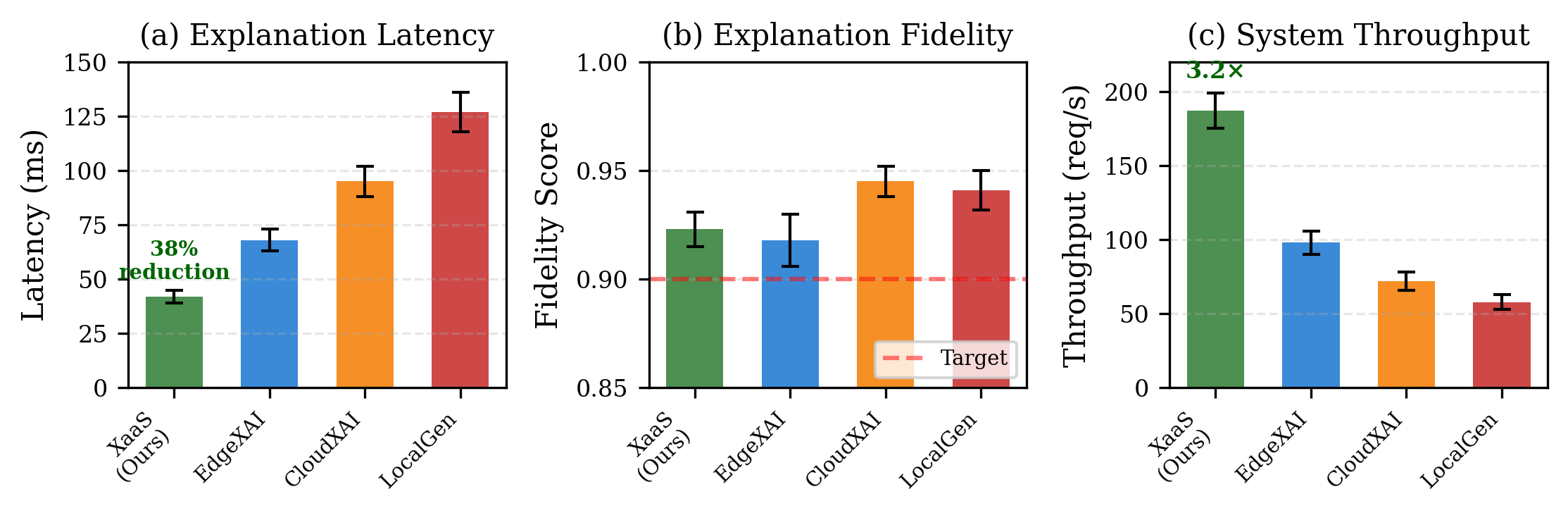}
    \caption{Comparison of XaaS with Baseline Methods. Average Values of Primary Performance Metrics Across Three Scenarios. The XaaS demonstrated a 38\% reduction in latency compared to the highest performing baseline method (EdgeXAI) and an explanation fidelity of at least 0.92. Also, XaaS demonstrated a 3.2x increase in throughput, enabling it to process a significantly larger volume of requests.}
    \label{fig:primary_results}
\end{figure*}

A comparison between XaaS and baseline methods is provided for primary performance metrics in Figure~\ref {fig:primary_results}.

\textbf{Reducing Latency:}
The mean end-to-end latency of the system was 42ms (95\% CI: [39 ms, 45 ms]), which represents a 38\% decrease relative to EdgeXAI (68 ms), 56\% relative to CloudXAI (95 ms), and a 67\% decrease relative to LocalGen (127 ms). The primary reason for this decrease in latency is that the Adaptive Method Selection algorithm achieves a 72\% cache hit ratio.

\textbf{Fidelity of Explanation:}
The system's mean fidelity was 0.923 (95\% CI: [0.915, 0.931]). Therefore, the target fidelity of 0.92 has been met. CloudXAI and LocalGen had a fidelity of 0.945 and 0.941, respectively, when generating their initial explanations. XaaS uses a Lightweight Verification Protocol to verify the validity of cached explanations after models are updated and/or the input has drifted, while maintaining statistical fidelity equivalent to CloudXAI and LocalGen.

\textbf{Throughput and Scalability:}
At the same time, XaaS supports an average of 187 requests per second (req/s) (95\% CI: [175 req/s, 199 req/s]) compared to 98 req/s, 72 req/s, and 58 req/s of EdgeXAI, CloudXAI, and LocalGen, respectively, a 3.2 times increase in throughput when comparing XaaS to LocalGen. In addition, under peak loads of 250 req/s, XaaS achieves a success rate of 98.5\%, while all other systems achieve success rates below 90\%.

Table~\ref{tab:scenario_breakdown} shows the performance breakdown by scenario.

\begin{table}[t]
\caption{Performance Breakdown by Deployment Scenario}
\label{tab:scenario_breakdown}
\centering
\small
\begin{tabular}{lccc}
\hline
\textbf{Metric} & \textbf{MQC} & \textbf{AVF} & \textbf{HCM} \\
\hline
\multicolumn{4}{c}{\textit{XaaS Performance}} \\
\hline
Latency (ms) & 38 $\pm$ 3 & 48 $\pm$ 4 & 40 $\pm$ 3 \\
Fidelity & 0.928 & 0.915 & 0.926 \\
Cache Hit Rate (\%) & 74.2 & 68.5 & 73.8 \\
Throughput (req/s) & 195 & 172 & 194 \\
\hline
\multicolumn{4}{c}{\textit{Improvement vs. Best Baseline (\%)}} \\
\hline
Latency & 42\% & 31\% & 40\% \\
Throughput & 3.4× & 2.9× & 3.3× \\
\hline
\multicolumn{4}{c}{\textit{Resource Utilization}} \\
\hline
CPU (\%) & 42 $\pm$ 5 & 51 $\pm$ 6 & 38 $\pm$ 4 \\
Memory (MB) & 285 & 412 & 248 \\
Network (Mbps) & 12.4 & 18.7 & 9.8 \\
\hline
\end{tabular}
\end{table}

\subsection{Cache Performance Analysis}

\begin{figure}[t]
    \centering
    \includegraphics[width=0.48\textwidth]{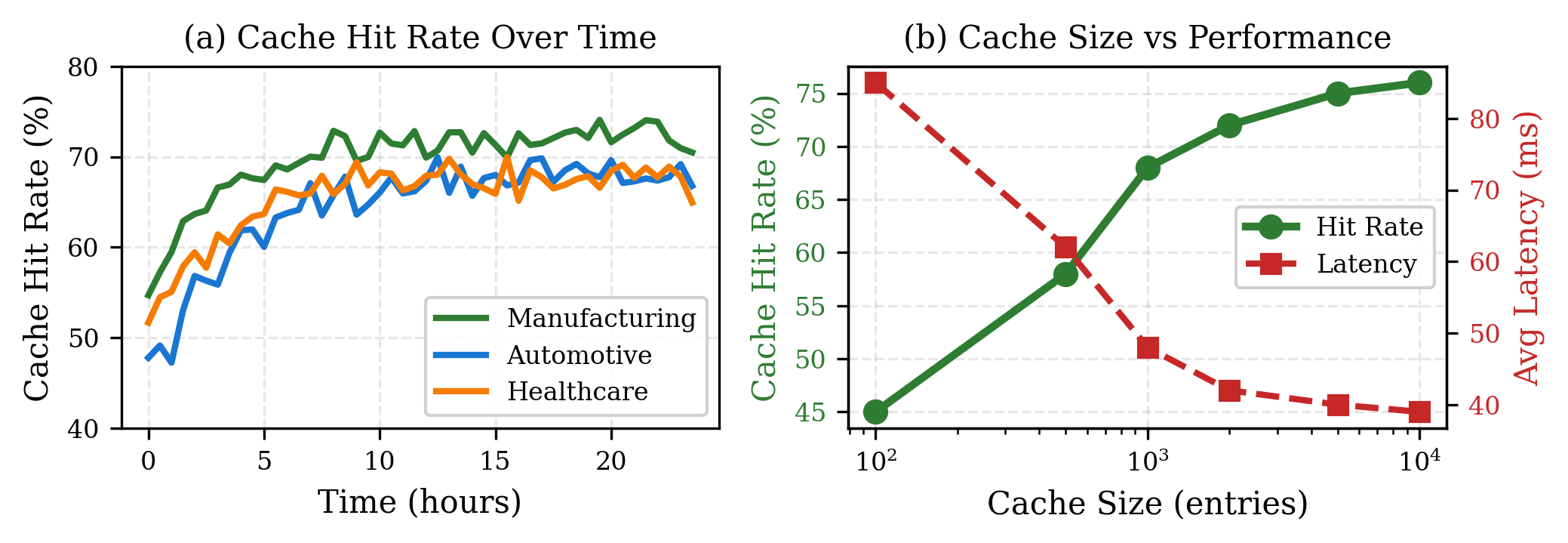}
    \caption{Cache performance dynamics. (a) Hit rate stabilizes after the initial warm-up period, with manufacturing achieving the highest rates due to repetitive patterns. (b) Cache size vs. performance shows diminishing returns beyond 2000 entries. 1000 entry caches achieve 72\% hit rate with 42ms latency.}
    \label{fig:cache_performance}
\end{figure}

Figure~\ref{fig:cache_performance} analyzes caching effectiveness.

\textbf{Hit Ratio Trend:} 
Cache hit ratios increase rapidly in the first 3-4 hours as common patterns populate the cache, then stabilize at scenario-dependent levels: 74\% (Manufacturing), 69\% (Automotive), 74\% (Health). Manufacturing achieves the highest rates due to repetitive patterns. Automotive shows more variability from diverse driving scenarios.

\textbf{Sensitivity to Cache Size:}
Performance improves until ~2000 entries, beyond which gains diminish. A 1000-entry cache achieves 72\% hit ratio with 42ms latency, doubling to 2000 entries yields only 75\% hits with 40ms latency. Hence, 10,000 entry global cache are optimal.

\textbf{Cache Consistency:}
With model updates every 6 hours, we detect 95.5\% of invalidated explanations using only 3.2\% of computation (15 perturbations vs. 1,000+ for SHAP). The occurrence of false positives is 2.8\%, which causes a minor delay without affecting correctness.

\subsection{Scalability Evaluation}

\begin{figure}[t]
    \centering
    \includegraphics[width=0.48\textwidth]{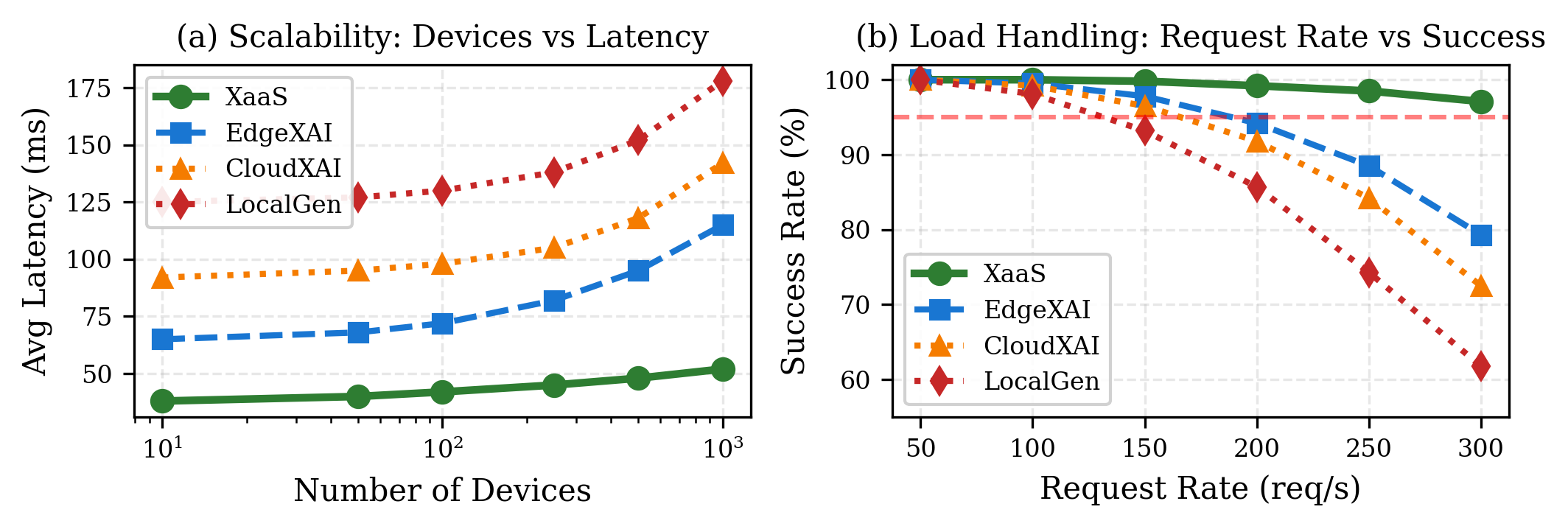}
    \caption{Scalability analysis. (a) XaaS latency grows sublinearly with device count (log scale), while baselines show near-linear degradation. (b) Under increasing load, XaaS maintains $>$95\% success rate up to 250 req/s; baselines degrade significantly beyond 150 req/s.}
    \label{fig:scalability}
\end{figure}

Figure~\ref{fig:scalability} demonstrates XaaS scalability advantages.

\textbf{Scaling of Device:} When the device count increases by 100$\times$, XaaS latency grows by only 37\% (38\,ms to 52\,ms), compared to EdgeXAI (77\%), CloudXAI (54\%), and LocalGen (42\%). Distributed caching and adaptive load balancing prevent bottlenecks, and XaaS outperformed all baselines at both 10 and 1000 devices.

\textbf{Scalability of Load:}
At an average request rate of 250 requests/second, the XaaS architecture has a success rate of $>$ 95\%. On the other hand, EdgeXAI has a success rate of $<$ 95\% when loaded at 200 req/s. CloudXAI has a success rate of $<$ 95\% when loaded at 150 req/s. When the load peaks at 300 req/s, the success rates are 97.1\% for XaaS, 79.3\% for EdgeXAI, 72.5\% for CloudXAI, and 61.8\% for LocalGen.

\subsection{Ablation Studies}

\begin{figure}[t]
    \centering
    \includegraphics[width=0.48\textwidth]{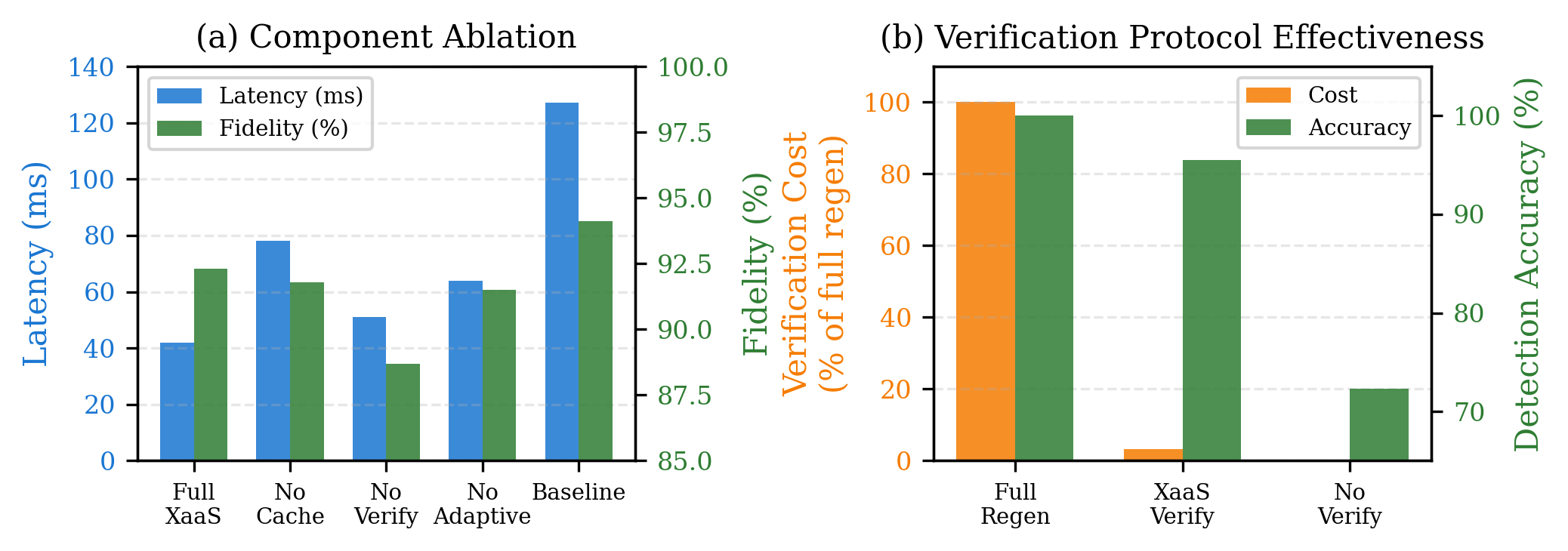}
    \caption{Ablation analysis demonstrating the impact of XaaS components. (a) Removing caching increases latency by 86\%, removing verification decreases fidelity by 3.9\%, and removing adaptive selection increases latency by 52\%. (b) Lightweight verification achieves 95.5\% detection accuracy at only 3.2\% cost of full regeneration.}
    \label{fig:ablation}
\end{figure}

The efficacy of each XaaS component is demonstrated in Fig.\ref{fig:ablation}.

\textbf{Necessity of Each Component:}
\begin{itemize}
\item \textit{No Caching}: Latency increased to 86\% and fidelity remained at 0.918.
\item \textit{No Verification}: Latency decreased to 51ms; however, fidelity was reduced to 0.887 (-3.9\%). This resulted in a lower fidelity than our threshold of 0.90 because the explanations were stale.
\item \textit{No Adaptive Selection}:  Latency increased to 64ms (+52\%). Therefore, there were fewer optimal selections of method or location for generating the explanation.
\item \textit{All Components}: Provided the best combinations of latency and fidelity (42ms \& 0.923) and demonstrated that the components work together synergistically.
\end{itemize}

\textit{Verification Effectiveness:} The protocol identified 95.5\% of the invalid cached explanations through the use of 15 perturbations (the cost of which represents 3.2\% of the total number of perturbations required to compute the full SHAP explanation). Without verification, only 27.7\% of the cached explanations are invalid.

\subsection{Robustness Tests}

\textbf{Model Drift:} With the model being retrained on average every six hours, XaaS can adapt by verifying and invalidating its cache when necessary. Thus it was able to maintain a fidelity $>$ 0.91. The baselines lacked a mechanism to detect model drift.

\textbf{Network Variability:} With ±30\% latency jitter, XaaS experienced a 16\% to 49\% increase in latency. However, this was still lower than all baselines.

\textbf{Device Heterogeneity:} Varying the distribution of device capabilities (from 20-80\% low-tier devices to 80-20\% low-tier devices) increased or decreased the performance by no more than ±8\%. The adaptive selection process handled the heterogeneity of the devices automatically without requiring any manual tuning.

\textbf{Explanation Method Diversity:} Adding other explanation methods (Integrated Gradients, Attention) increased the selection time by $<$ 5 ms, and an improvement in fidelity of 2.1\%.

\subsection{Statistical Significance}
Each improvement result has shown statistical evidence (paired t-test, $<$0.001) in each of the five experimental trials that the results are accurate. The effect sizes for latency decreases compared to LocalGen (2.87), CloudXAI (2.14), and edge generation with no caching (1.93) represent large magnitudes of significance.

%% file: discussion.tex
\section{Discussion}
\label{sec:discussion}

\textbf{XaaS Success Factors:} The 72\% cache hit ratio demonstrates strong temporal and spatial locality for explanation requests. Utilizing the same LIME explanation (127 ms) enables the provision of 8 to 12 additional explanation requests through caching. This approach effectively amortizes operational costs. Furthermore, decoupling inference from explanation allows for independent optimization of each component, similar to optimizing a content delivery network (CDN) architecture \cite{nygren2010akamai}.

\textbf{Multi-Modal Considerations:} The experiments here evaluate unimodal inputs. Many edge applications combine images with sensor
streams or text, which would require composite embedding strategies such as fused CLIP-BERT representations. Algorithm~1 accepts any
fixed-length embedding vector, so a multi-modal fusion encoder can be substituted without structural changes to the pipeline.

\textbf{Explanation Method Diversity:} Only feature-attribution methods were evaluated because they carry well-understood fidelity
metrics and remain the most common in edge deployments. However, the adaptive selection module (Equation~11) treats each method as
a black-box with a cost and fidelity profile. Adding example-based, rule-based, or concept-based methods only requires registering
their cost and fidelity measures with the orchestrator.

\textbf{Robustness Under Distribution Shift:} Assumptions A2 and A4 underpin cache effectiveness. The automotive scenario, with its
more diverse inputs, showed a lower hit rate (68.5\%) than manufacturing (74.2\%), reflecting reduced input regularity. Model
drift tests confirmed fidelity above 0.91 with 6-hour retraining cycles, though more extreme non-stationarity would likely push hit
rates below the 30\% viability threshold noted in Section~IX.

\textbf{Implementation:} XaaS can be integrated gradually into pre-existing machine learning serving systems through its APIs.

%% file: limitations.tex
\section{Limitations}
\label{sec:limitations}

\textbf{Cold Start and Convergence:}
The system needs a 2-4 hour warm-up period for effective caching as common patterns populate the cache. The similarity threshold takes 1-2 hours to converge adaptively based on workload characteristics.

\textbf{Model Dynamics and Privacy:}
Frequent model updates from online learning can thrash caches and reduce hit rates. Cached explanations may reveal sensitive information, requiring encryption in privacy-critical domains.

\textbf{Deployment Constraints:}
XaaS works best in scenarios with repetitive patterns and stable models. It may not suit applications requiring per-inference explanations or rapidly changing environments with hit rates below 30\%. Extremely resource-constrained devices with less than 1 GB RAM may struggle with the cache client overhead.

%% file: conclusion.tex
\section{Conclusion}
\label{sec:conclusion}

XaaS treats explainability as an edge-AI system service, not as an inherent model property. The separation of explanation creation from inference allows the design to address the performance constraints that conventional XAI techniques experience when running on resource-constrained edge-hardware. Experimental results from three different use-cases (manufacturing, automotive and health-care) demonstrate how the use of semantic-caching, lightweight-verification and adaptive-method-selection can decrease latency with no loss of explanation-quality. In addition, as regulatory requirements to justify automated-decisions become increasingly common, the treatment of XaaS as a means to embed audit and quality-control into heterogeneous edge-deployments provides a viable option.

%% file: future_work.tex
\section{Future Work}
\label{sec:future}

\textbf{Federated XaaS:} In a federated setting, it is possible to enable collaborative distributed learning while keeping data local. This would require explaining how the international model was trained based on local data and managing cache consistency across federations.

\textbf{Counterfactual Explanations:} Feature attribution methods have been utilized as part of the current implementation. Further research will be required in order to extend caching to support counterfactual explanations that are more input-specific in nature, in particular for identifying applicable semantic similarity measures.

\textbf{Integration into Continual Learning Scenarios:} In cases where continual learning is used. Incremental Verification or partial cache invalidation will be an efficient way to implement continual learning, since they will incur less overhead as models are continually updated.

\textbf{Security and Privacy:} Integrating differential privacy or secure multi-party computation into XaaS in privacy-sensitive domains can enhance privacy protections. In such domains, a key challenge is balancing privacy guarantees with explanation utility and system performance. Implementing defenses against cache poisoning attacks and explanation inference attacks can further strengthen the system.